%% file: main.tex
\documentclass[11pt, a4paper, goog]{google}

\usepackage[T1]{fontenc}
\usepackage[utf8]{inputenc}
\usepackage[authoryear, sort&compress, round]{natbib}
\usepackage{multirow}
\usepackage{array}
\usepackage{placeins}
\usepackage{tikz}
\usetikzlibrary{arrows.meta, positioning, fit, backgrounds, calc, shapes.geometric}
\usepackage[most]{tcolorbox}
\usepackage{algorithm}
\usepackage{algpseudocode}
\usepackage{float}
\usepackage{wrapfig}
\newcommand{\apptocline}[1]{\noindent\makebox[1.8em][l]{\ref{#1}}\nameref{#1}\dotfill\pageref{#1}\par\vspace{3pt}}

\theoremstyle{remark}

\newcommand{\harnessname}{\textsc{VeriHarness}}
\definecolor{gainpos}{RGB}{19,115,51}
\definecolor{gainneg}{RGB}{170,35,35}
\definecolor{oursbg}{HTML}{E6F4EA}

\uselogo{}

\title{\harnessname{}: Scaling Agentic Verification for Long-Horizon Tasks}

\correspondingauthor{cz391@cam.ac.uk, \{rujunh, chenyulee\}@google.com}
\worknote{* This work was done while Caiqi interned at Google Cloud AI Research.}

\author[1*]{Caiqi Zhang}
\author[2]{Rujun Han}
\author[2]{Zifeng Wang}
\author[2]{Zoey CuiZhu}
\author[1]{Nigel Collier}
\author[2]{Tomas Pfister}
\author[2]{Chen-Yu Lee}
\affil[1]{University of Cambridge}
\affil[2]{Google Cloud AI Research}

\begin{abstract}
As LLM agents undertake increasingly complex, long-horizon tasks, verifying their outputs becomes increasingly challenging.
We study how verification capability can be strengthened with a fixed base model, without access to reference answers or grading rubrics at test time.
Repeated sampling yields multiple rollouts that can contain complementary correct claims, but we need a reliable verification mechanism to determine which claims to trust.
We first find that disagreement often exposes correct alternatives, while consensus can conceal errors. These observations motivate \harnessname{}, which turns the underlying LLM a generator uses into an agentic verifier by giving it a workspace, evidence tools, and reusable verification skills.
A disagreement resolver checks competing claims against environmental evidence, while a consensus challenger tests shared claims and searches for omitted requirements.
Their findings guide the selection and revision of the final artifact.
Across five long-horizon workspace benchmarks and two frontier models, \harnessname{} achieves the highest selection scores among the evaluated baselines. Evidence-backed revision further improves average performance, bringing gains over a single rollout to 6.2 points with Gemini~3.5~Flash and 6.4 points with Claude~Opus~4.8.
We further show that verification skills can self-improve from failure feedback, demonstrating \harnessname{} as a novel and critical approach for scaling long-horizon agentic verification.
We release the full pool of approximately 26,000 rollouts across all five benchmarks and both models, produced at a cost of over \$100{,}000, to support future research on agentic verification.

\vspace{4pt}
{\centering\small
\raisebox{-0.16\height}{\includegraphics[height=1.05em]{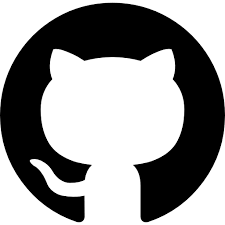}}\hspace{0.32em}\href{https://github.com/google-research/veriharness}{Code}%
\hspace{1.15em}\textcolor{gray!45}{\rule[0.02em]{0.6pt}{0.9em}}\hspace{1.15em}%
\raisebox{-0.16\height}{\includegraphics[height=1.05em]{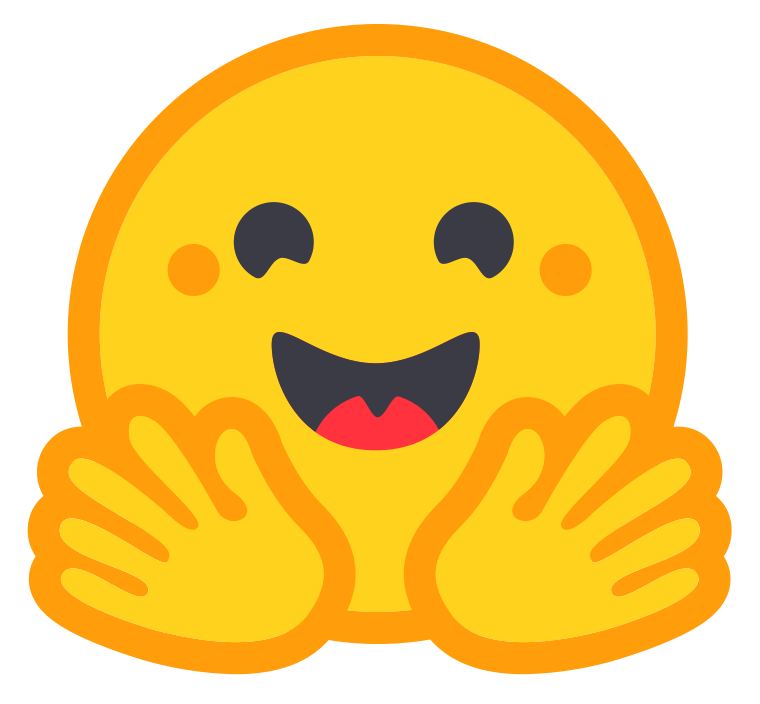}}\hspace{0.32em}\href{https://huggingface.co/datasets/caiqizh/veriharness}{Dataset}%
\hspace{1.15em}\textcolor{gray!45}{\rule[0.02em]{0.6pt}{0.9em}}\hspace{1.15em}%
\raisebox{-0.16\height}{\includegraphics[height=1.05em]{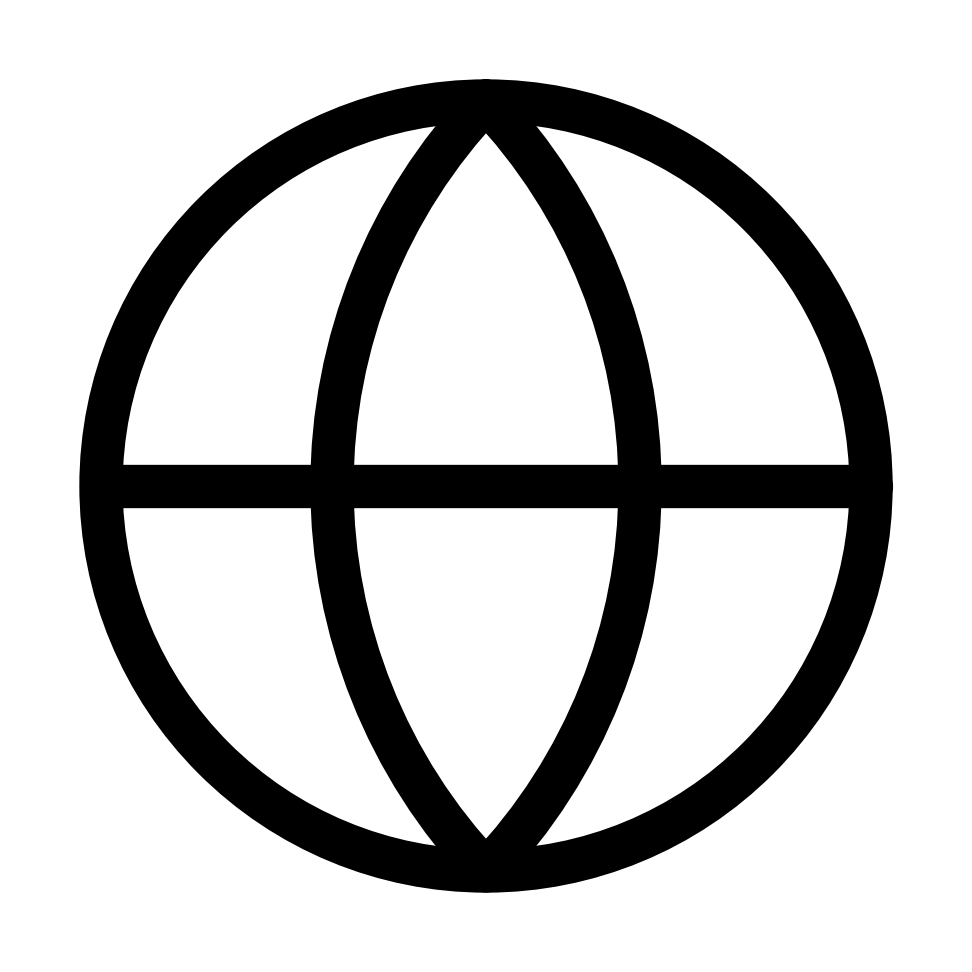}}\hspace{0.32em}\href{https://veriharness.com}{\texttt{veriharness.com}}%
\par}
\end{abstract}

\begin{document}

\maketitle

\begin{figure}[H]
\centering
\includegraphics[height=1.62in]{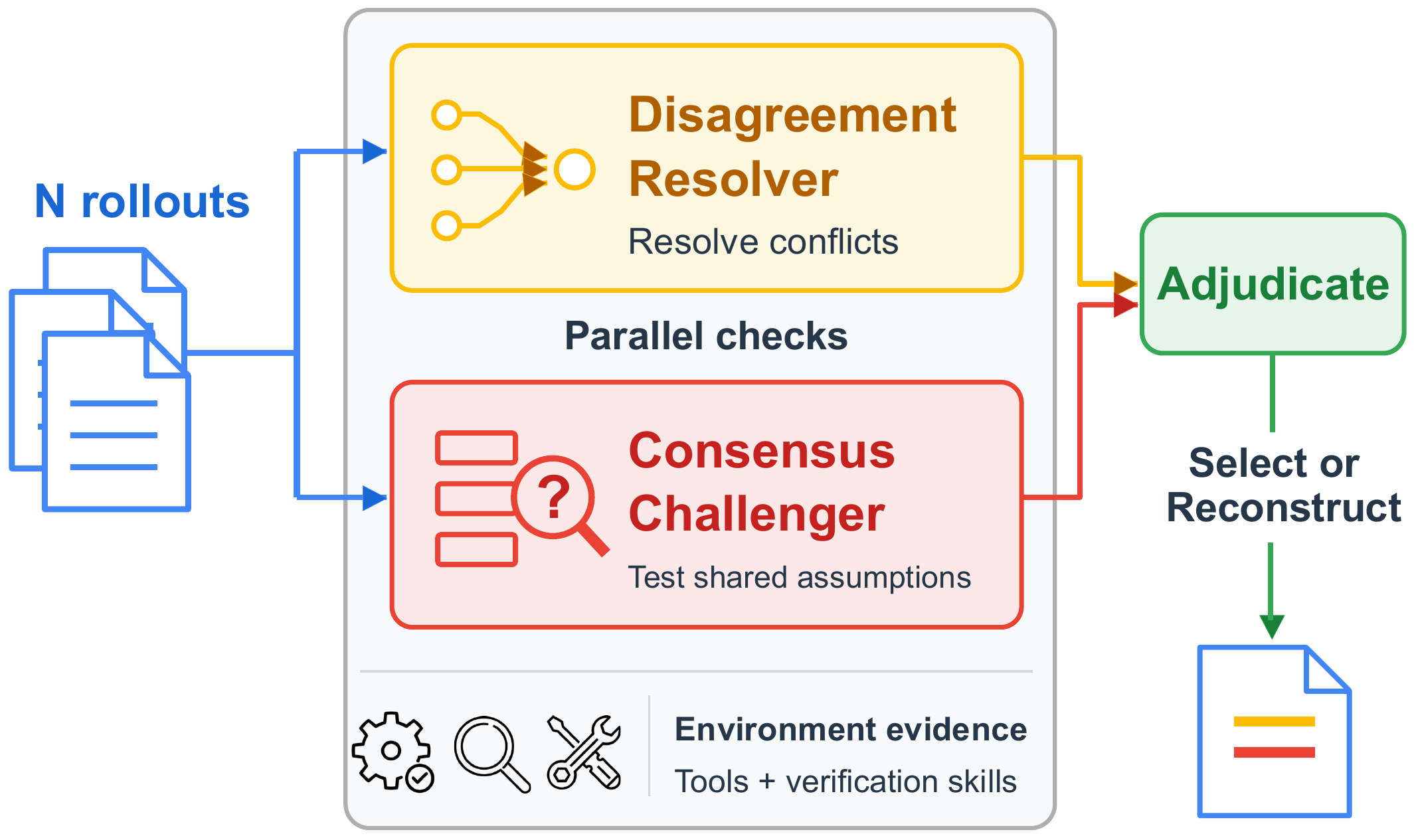}\hfill
\includegraphics[height=1.62in]{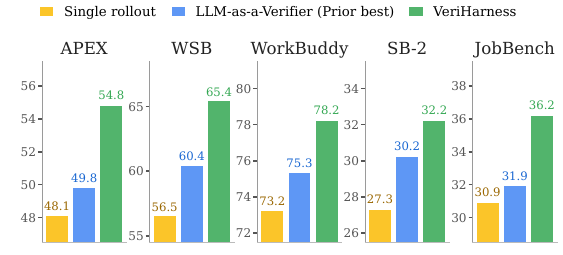}
\caption{\textbf{\harnessname{} and its results.} \textit{Left:} the verifier is the generator's own model inside a harness. A disagreement resolver tests the claims on which the $N$ rollouts differ, and a consensus challenger seeks evidence against the claims they share. Adjudication combines their findings to guide artifact selection and revision. \textit{Right:} with Gemini~3.5~Flash as both generator and verifier, \harnessname{} with revision improves over the single-rollout baseline on all five benchmarks, including gains of \textbf{6.7 points} on APEX-Agents and \textbf{8.9 points} on Workspace-Bench Lite (WSB).}
\label{fig:summary}
\end{figure}

\input{sections/01-introduction}
\input{sections/02-problem}
\input{sections/03-method}
\input{sections/05-experiments}
\input{sections/06-self-evolving}
\input{sections/07-analysis}
\input{sections/08-related-work}
\input{sections/09-conclusion}
\input{sections/10-limitations}

\bibliography{reference}

\clearpage
\appendix

\section*{Appendix Contents}
\apptocline{app:claim-analysis}
\apptocline{app:same-model-verification}
\apptocline{app:procedure}
\apptocline{app:experimental-details}
\apptocline{app:release}
\apptocline{app:ablation}
\apptocline{app:runtime}
\apptocline{app:tools-skills}
\apptocline{app:evolve}
\apptocline{app:challenger}
\apptocline{app:cost}
\vspace{1.5em}

\input{sections/appendix}

\end{document}

%% file: sections/01-introduction.tex
\section{Introduction}
\label{sec:introduction}

Long-horizon agents increasingly produce reports, spreadsheets, and other artifacts whose correctness requires substantial work to assess~\citep{apexagents2026, workspacebench2026, workbuddy2026}. Recent workspace benchmarks show that even frontier models struggle to produce reliable artifacts~\citep{spreadsheetbench2, jobbench2026}. Verifying these artifacts is harder still. In coding and mathematics, unit tests, executable checks, or formal proof checkers give direct feedback on correctness even when no reference answer exists~\citep{ehrlich2025codemonkeys, li2025sstar}. General complex tasks rarely offer an equally straightforward way to assess the entire output. For example, checking a report may require tracing claims to sources, recomputing quantities, and interpreting requirements across several files. As agents take on more of this work, verification becomes a central challenge and must develop alongside generation~\citep{you2026agentasajudge}.

We study test-time verification without a stronger judge: the generator and verifier share the same model and tools, and the verifier checks the output artifacts without access to reference answers or grading rubrics. This setting reflects practice at the frontier, where the generator is already the strongest available model and no stronger judge exists to oversee it~\citep{bowman2022scalableoversight, burns2024weaktostrong}. Holding the base model fixed also ensures that no knowledge from a stronger model enters the verifier: any gain comes from the model's own rollouts and environment~\citep{huang2023selfimprove}. In this setting, the verifier draws on two unique sources of information: 1) the candidate rollouts and 2) the task environment. Repeated sampling yields several rollouts of the same task~\citep{brown2024monkeys, scalingttcagents2025}, and comparing them reveals competing claims and shared conclusions. The task environment provides source files, data, and constraints against which those claims can be checked. 

\begin{wrapfigure}{r}{0.42\linewidth}
\vspace{-10pt}
\centering
\includegraphics[width=\linewidth]{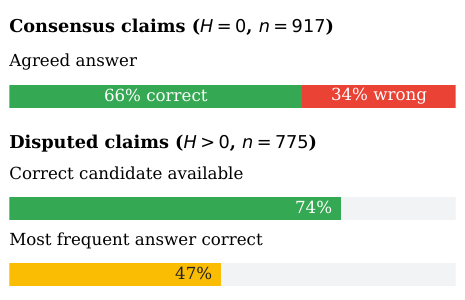}
\caption{\textbf{Consensus can preserve errors; disagreement can contain correct alternatives that voting discards.} APEX-Agents claims from ten-rollout pools of Claude~Opus~4.8; percentages are within each group, correctness uses the benchmark's rubric grades, and $H$ is the claim entropy of Section~\ref{sec:observation}.}
\label{fig:claims}
\vspace{-8pt}
\end{wrapfigure}

Existing test-time verification methods mostly use only the rollouts. Majority voting and self-consistency~\citep{wang2023selfconsistency, chen2023usc} treat the most common answer as correct, and LLM-as-a-judge approaches~\citep{zheng2023judge, llmasverifier2026} score rollouts by reading them. Both rely on what the model already believes rather than on evidence the environment could provide. To address this shortcoming, we first propose to bring verification into the environment: individual claims are tested against the workspace's source data and constraints. Second, we find that \textbf{consensus does not guarantee correctness, while disagreement between rollouts can expose correct alternatives.} A case study in Figure~\ref{fig:claims} shows that roughly a third of the agreed values are judged incorrect, so errors cannot be corrected by voting or selection alone. Also, disputed claims often include a correct candidate, although the most frequent answer may be wrong (Section~\ref{sec:observation}). Checking these alternatives against evidence can help identify the correct value. Therefore, a verifier has to test competing answers and challenge the consensus that all candidates share.

We introduce \harnessname{} (Figure~\ref{fig:summary}), a general-purpose, plug-and-play harness for the verifier, which uses the same underlying model as the generator. The harness encompasses a workspace, evidence tools, and reusable verification skills to support these two checking tasks. A \emph{disagreement resolver} tests competing claims against source files, data, and task constraints. A \emph{consensus challenger} searches for evidence that could refute consensus values or reveal omitted requirements. Their findings guide the selection and revision of the final artifact, accompanied by a record of the supporting evidence. The verifier decides what to check, how to interpret the evidence, and which changes to make.
We further ask whether the harness can grow its verification skills with experience. Building on reusable skill and experience libraries~\citep{wang2023voyager, ouyang2025reasoningbank}, we let the harness accumulate skills from failure feedback on development tasks, starting from either an empty library or a human-authored one.

We evaluate on five demanding workspace benchmarks with two frontier models, Gemini~3.5~Flash and Claude~Opus~4.8, each serving as both generator and verifier. \harnessname{} achieves the highest selection score among the evaluated baselines on every benchmark with both models. Evidence-backed revision further improves average performance, yielding gains over a single rollout of 6.2 points with Flash and 6.4 with Opus (Table~\ref{tab:main}). On held-out tasks, skills evolved from an empty library outperform the human-authored library, and evolving from the human-authored library adds 6.8 points on APEX-Agents and 3.7 on SpreadsheetBench~2 over that library (Section~\ref{sec:self-evolving}).

Our contributions are threefold:
\begin{itemize}[leftmargin=*, itemsep=2pt]
\item \textbf{Verification principle.} We identify two complementary checking tasks in same-model verification: resolving disputed claims and challenging consensus claims against evidence from the environment.
\item \textbf{General-purpose harness.} We build \harnessname{}, the first agentic verification harness for long-horizon tasks, which is training-free and plug-and-play across benchmarks and models, and evaluate its ability to select and revise agent outputs.
\item \textbf{Evolving verification skills.} We show that a fixed model can accumulate useful verification skills from failure feedback, with human expertise providing an effective starting point for evolution.
\end{itemize}

%% file: sections/02-problem.tex
\section{Problem Formulation}
\label{sec:problem}
\label{sec:observation}
\noindent\textbf{Task and rollouts.} A task $t=(s,\mathcal{E})$ consists of a description $s$ and an environment $\mathcal{E}$ containing source files, data, and application state. The \emph{generator} is an agent built on model $M$ that attempts the task. Each attempt produces a \emph{rollout} $r_i=(d_i,w_i,\tau_i)$, comprising a delivered artifact $d_i$, final workspace state $w_i$, and action trace $\tau_i$, and repeated sampling yields a pool $R=\{r_1,\dots,r_N\}$. An artifact makes \emph{claims}: assertions about task-relevant values, interpretations, or the satisfaction of requirements, such as a reported figure, the reading of a clause, or a required section.

\noindent\textbf{Verification objective.} The \emph{verifier} uses the same model and tools as the generator. Given the task and the pool, it delivers a rollout $r^\star$, selected from $R$ or revised from one of its members, together with a record $C$ of checks, findings, and unresolved issues. Its objective is to maximize the quality of $r^\star$ as judged by an external grader. The verifier has no access to this grader, to reference answers, or to grading rubrics, and all compared methods receive the same rollout pool.

\noindent\textbf{Checks and evidence.} We examine artifacts at the level of claims. A \emph{check} $\chi$ is an operation that tests a claim, such as reading a source passage, recomputing a total, or inspecting file metadata, and its result $e=\mathcal{E}(\chi)$ is the evidence obtained by running the check in the environment. For example, ``FY2025 revenue is EUR120m'' is a claim, and inspecting the final report for the amount and currency is a check. Evidence acquisition is \emph{active}: the verifier chooses each check from the task, the candidate pool, and the evidence gathered so far. An informative check helps distinguish candidate answers or test a constraint that the current answer may violate. Evidence can support a candidate, contradict it, or establish a correction absent from the pool, and a claim remains unresolved when the evidence is insufficient for a verdict. Task requirements also identify claims to examine when the corresponding content is absent from every artifact.

\noindent\textbf{Consensus and disputed claims.} Let $p$ identify a common question or requirement, and let $v_{i,p}$ denote the answer or content supplied by artifact $d_i$, treating equivalent expressions as the same value and omission as a distinct value. The candidate set $\mathcal{V}_p=\{v_{i,p}\}_{i=1}^{N}$ has empirical distribution and entropy
\begin{equation}
\pi_p(v)=\frac{1}{N}\sum_{i=1}^{N}\mathbf{1}[v_{i,p}=v],\qquad
H(\pi_p)=-\sum_{v\in\mathcal{V}_p}\pi_p(v)\log\pi_p(v).
\label{eq:claim-entropy}
\end{equation}
A claim is a \emph{consensus} claim when $H(\pi_p)=0$ and a \emph{disputed} claim when $H(\pi_p)>0$. The entropy measures variation across candidates only: unanimity can hold even when every candidate is incorrect.

\noindent\textbf{Empirical observations.} On ten-rollout pools of Claude~Opus~4.8 on APEX-Agents, whose rubrics permit claim-level grading (details in Appendix~\ref{app:claim-analysis}), 34\% of consensus values are judged incorrect, while 74\% of disputed claims contain a correct candidate and the most frequent value is correct in only 47\% (Figure~\ref{fig:claims}). These statistics motivate two complementary checking tasks. For disputed claims, the verifier seeks evidence that distinguishes the exposed alternatives, while allowing that every candidate may be incorrect. For consensus claims, it must first propose how the consensus value could be wrong and then test those proposals against the environment. Proposing failures requires knowledge of how artifacts fail, which the pool does not supply. Section~\ref{sec:method} develops a harness for both tasks.

%% file: sections/03-method.tex
\section{\harnessname{}}
\label{sec:method}

\begin{figure}[t]
\centering
\includegraphics[width=\linewidth]{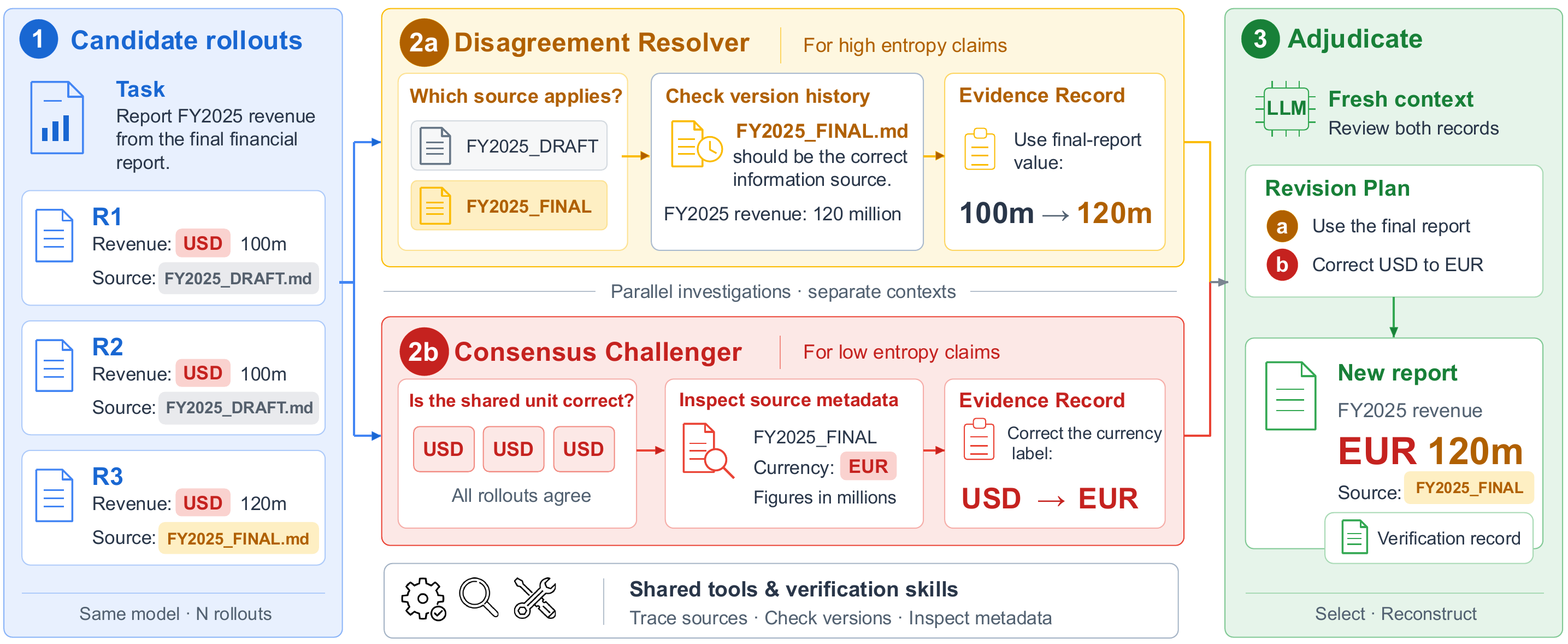}
\caption{\textbf{\harnessname{} on a worked example.} The task asks for FY2025 revenue from the final financial report. (1) Two rollouts cite the draft and report 100m; a third cites the final report and gives 120m, and all three label the figure USD. (2a) The resolver checks version history and establishes that the final report supersedes the draft, supporting 120m. (2b) In a separate context, the challenger checks source metadata and finds that the currency every rollout uses should be EUR. (3) Adjudication in a fresh context combines the evidence into a revision plan, and delivery produces the revised report with its verification record.}
\label{fig:overview}
\end{figure}

\harnessname{} implements the two checking tasks of Section~\ref{sec:observation} with the same model as the generator. As Figure~\ref{fig:overview} illustrates, a disagreement resolver and a consensus challenger use shared tools and verification skills to produce evidence records. Adjudication in a fresh context turns both records into a selection and revision plan for the delivered artifact. Appendix~\ref{app:same-model-verification} discusses why the same model can detect errors made during generation.

\subsection{A General-Purpose Verification Harness}
\label{sec:harness}

An agent harness $\mathcal{H}_G$ supplies the tools, memory, and operating loop that turn a frozen model $M$ into the generator, $\mathrm{Generator}=(M,\mathcal{H}_G)$. \harnessname{} applies the same construction to verification:
\begin{equation}
\mathrm{Verifier}=(M,\mathcal{H}_V),\qquad \mathcal{H}_V=(\mathcal{W},\mathcal{T},\Pi,\mathcal{S}),
\label{eq:harness}
\end{equation}
where $\mathcal{W}$ is the workspace, $\mathcal{T}$ the evidence tools, $\Pi$ the verification protocol, and $\mathcal{S}$ the skill library. Table~\ref{tab:harness} summarizes how they provide access, organize checking, and supply reusable knowledge.

\noindent\textbf{Workspace and tools.} The workspace exposes each rollout's artifact $d_i$, final state $w_i$, and trace $\tau_i$ as files alongside the task description $s$ and environment $\mathcal{E}$. The verifier can trace a claim to its source or computation, recompute quantities, and execute code to obtain evidence $\mathcal{E}(\chi)$ for a check $\chi$.

\noindent\textbf{Verification protocol.} The protocol $\Pi$ assigns the two checking tasks to separate model contexts, followed by adjudication and delivery. The program provides the contexts and information access; the model chooses which claims to examine, which checks to run, and when to stop. Because the workspace is file-based and the protocol does not depend on the artifact type, the same harness applies to reports, spreadsheets, and code; only the skills hold artifact-specific knowledge.

\noindent\textbf{Verification skills.} A skill describes a reusable failure mode and its checking procedure in a short text with an optional script. Tools perform a check, while skills identify what needs checking: a tool reads a financial table, whereas a skill directs the verifier to test whether a growth rate spans the years in its label. Skills exclude task-specific answers. Section~\ref{sec:experiments} uses a human-authored library $\mathcal{S}$, listed in full with the tools in Appendix~\ref{app:tools-skills}; Section~\ref{sec:self-evolving} studies how failure feedback can extend this library without changing the model or protocol.

\subsection{Resolving Disagreement through Evidence}
\label{sec:resolver}

The disagreement resolver compares artifacts to locate a disputed claim $p$ and candidate values $\mathcal{V}_p$, which it traces to their sources and computations. It chooses the check that best distinguishes them. Verification skills guide these choices, for example the source-version check in Figure~\ref{fig:overview} and checks of competing interpretations of a workbook convention. Let $b_p$ be its current belief over the correct value given the evidence $e_{<k}$ gathered before step $k$, distinct from the empirical frequencies $\pi_p$ in Section~\ref{sec:observation}. The resolver prefers checks with large expected information gain~\citep{lindley1956},
\begin{equation}
\mathrm{Gain}(\chi)=\mathbb{E}\big[\,H(b_p)-H(b_p(\cdot\mid \mathcal{E}(\chi)))\ \big|\ e_{<k}\big],
\label{eq:voi}
\end{equation}
which it judges qualitatively rather than computes. After running a check and obtaining evidence $e$, the resolver eliminates contradicted candidates:
\begin{equation}
\mathcal{V}_p\leftarrow\{v\in\mathcal{V}_p:\ v\ \text{is consistent with}\ e\}.
\label{eq:eliminate}
\end{equation}
The resolver closes a claim when the evidence settles it and stops when it judges that further checks are unlikely to change the outcome. If the evidence contradicts every candidate, it records a new value only when the evidence establishes one; otherwise the claim remains open. The resolver records all outcomes in the evidence record $L_{\neq}$ for joint adjudication with the challenger's findings.

\subsection{Challenging Consensus}
\label{sec:challenger}

Consensus supplies no competing value to test, so the challenger identifies consensus claims, proposes how each could fail, and tests those possibilities against the environment. It prioritizes checks by how likely they are to expose a task-relevant error given the task, the pool, and the evidence so far,
\begin{equation}
\mathrm{Priority}(\chi)=P\big(\chi\ \text{exposes a failure of the consensus claim}\ \big|\ t,R,e_{<k}\big),
\label{eq:falsify}
\end{equation}
which it judges qualitatively rather than estimates numerically. The challenger examines three kinds of consensus claim:
\begin{itemize}[leftmargin=*, itemsep=2pt]
\item \textbf{Consensus values.} The challenger recomputes quantities from raw inputs and compares labels with source metadata, including the currency check in Figure~\ref{fig:overview}.
\item \textbf{Consensus readings.} The challenger tests a common interpretation against the file the task names, the period the source states, or the definition the workbook carries.
\item \textbf{Consensus omissions.} The challenger checks the delivered artifacts against the task description $s$ for requirements that every rollout overlooked.
\end{itemize}
An omission must first be made visible as a verification target because it appears in no artifact. More generally, challenging consensus requires knowledge of how an apparently settled artifact can fail. Skills supply that knowledge: a sign-convention skill, for example, gives the challenger a concrete property to test despite unanimous values. The challenger records its findings in the evidence record $L_{=}$.

\subsection{Adjudication and Evidence-Backed Revision}
\label{sec:adjudication}

\noindent\textbf{Adjudication.} The investigations run in separate contexts so that disputed claims do not displace attention from consensus claims; Appendix~\ref{app:ablation} evaluates this separation. A fresh context of the same model then reviews both records with the task and rollouts, without inheriting either investigator's conversational history. It can uphold, overturn, or leave findings unresolved. It outputs a base rollout, an evidence-supported revision plan, and a list of unresolved claims:
\begin{equation}
(\mathrm{base},\ \mathrm{revision\_plan},\ \mathrm{unresolved})=A(L_{\neq},L_{=},t,R),
\label{eq:adjudicate}
\end{equation}
where $\mathrm{base}\in R\cup\{\varnothing\}$ is the rollout to build on, $\mathrm{revision\_plan}$ specifies the changes to make and their supporting evidence, and $\mathrm{unresolved}$ lists claims that the evidence has not settled. Adjudication selects the base with the fewest evidence-supported defects on the task's substantive requirements, weighing findings by their importance to the requested deliverable. When no candidate provides a usable starting point, it sets $\mathrm{base}=\varnothing$.

\noindent\textbf{Delivery and verification record.} In selection-only evaluation, the adjudicated $\mathrm{base}\in R$ is returned unchanged. The full harness applies the revision plan and returns the resulting rollout with its verification record:
\begin{equation}
\begin{aligned}
r^\star&=\mathrm{Apply}(\mathrm{base},\ \mathrm{revision\_plan},\ \mathrm{unresolved}),\\
C&=(L_{\neq},\ L_{=},\ \mathrm{base},\ \mathrm{revision\_plan},\ \mathrm{unresolved}).
\end{aligned}
\label{eq:delivery}
\end{equation}
An empty revision plan leaves the base unchanged; a nonempty plan revises it, or produces a new artifact from the plan when $\mathrm{base}=\varnothing$. Claims in $\mathrm{unresolved}$ stay marked as unsettled in $C$, and delivery presents both readings where the artifact format allows.

Each evidence record links a claim, check, evidence, and verdict, making changes traceable and preserving open issues. Together with grader feedback on development tasks, these records support skill development in Section~\ref{sec:self-evolving}. Appendix~\ref{app:procedure} gives the complete procedure and delivery rules.

%% file: sections/05-experiments.tex
\section{Experiments}
\label{sec:experiments}

\label{sec:setup}
\subsection{Setup}
\noindent\textbf{Evaluation scope.} We evaluate on five recent long-horizon workspace benchmarks spanning professional documents, spreadsheets, code, and multi-file tasks: APEX-Agents (version 1.0)~\citep{apexagents2026}, Workspace-Bench Lite~\citep{workspacebench2026}, WorkBuddy Bench~\citep{workbuddy2026}, SpreadsheetBench~2~\citep{spreadsheetbench2}, and JobBench~\citep{jobbench2026}. Table~\ref{tab:benchmarks} summarizes their scale, artifacts, and graders. Appendix~\ref{app:experimental-details} provides further benchmark details.

\begin{table}[htbp]
\centering
\caption{\textbf{Benchmarks.} Tasks is the number of tasks in our pools; every task has $N=10$ rollouts per model.}
\label{tab:benchmarks}
\footnotesize
\setlength{\tabcolsep}{4pt}
\begin{tabular}{l r l l l}
\toprule
Benchmark & Tasks & Domain & Artifact & Grader \\
\midrule
APEX-Agents (v1.0) & 480 & banking, law, consulting & memos, models, answers & expert rubric \\
Workspace-Bench Lite & 100 & file-heavy workspaces & documents and files & rubric judge \\
WorkBuddy Bench & 200 & code, office, web & patches and office files & composite verifier \\
SpreadsheetBench~2 & 321 & business spreadsheets & workbooks & recompute, compare \\
JobBench & 65 & 35 white-collar occupations & multi-file artifacts & rubric judge \\
\bottomrule
\end{tabular}
\end{table}

\noindent\textbf{Same-model comparison.} We use Gemini~3.5~Flash and Claude~Opus~4.8 as both generator and verifier, with thinking level \textit{high}. For every task and model, all methods receive the same frozen pool of $N=10$ rollouts and the same model. Outputs are scored by the benchmark graders, which are never available to the verifier. Appendix~\ref{app:experimental-details} gives the task, runtime, and grading details.

\label{sec:baselines}

\noindent\textbf{Baselines.} We compare \harnessname{} with majority voting, best-of-$N$ judging, a pairwise tournament, and the most recent LLM-as-a-Verifier~\citep{llmasverifier2026}. We also implement an \emph{agentic verifier (env.\ access)} given the same workspace and evidence tools as our verifier but no VeriHarness protocol or skills. \harnessname{} (select) returns the base chosen by adjudication, and \harnessname{} also applies the evidence-backed revision plan. For the latter, we add aggregation of the pool into a new artifact~\citep{wang2024moa, aggagent2026}, which also produces an artifact absent from the pool. We also include an \emph{agentic verifier + revision} baseline: the same model checks the pool against the workspace, selects a base, and revises it using the same evidence tools. The grader-informed selection oracle bounds selection from the fixed pool. We report each benchmark's native score over three independent verifier runs. Appendix~\ref{app:experimental-details} defines the baselines, metrics, and revision protocol.

\label{sec:results}

\begin{table}[!b]
\centering
\caption{\textbf{Main results.} Three-seed mean score on each benchmark's own scale, with the cross-seed standard deviation in gray; Avg is the unweighted mean over the five benchmarks, and gains are over the single rollout. Within each model, bold marks the best score in the primary selection comparison and, separately, among the methods with revision; CLI variants are reported separately, followed by methods that produce new artifacts. The single rollout is a reference baseline. The selection oracle takes the highest-scoring rollout per task under the benchmark grader and bounds selection from the fixed pool; it does not bound revision. SB-2: SpreadsheetBench~2; WSB: Workspace-Bench Lite.}
\label{tab:main}
\footnotesize
\setlength{\tabcolsep}{4.5pt}
\renewcommand{\arraystretch}{1.12}
\newcommand{\cs}[2]{\makebox[1.8em][r]{#1}\makebox[2.0em][l]{\scriptsize\color{gray}$\,\pm#2$}}
\newcommand{\cm}[1]{\makebox[1.8em][r]{#1}\makebox[2.0em][l]{}}
\newcommand{\cg}[1]{\makebox[1.8em][r]{#1}\makebox[2.0em][l]{}}
\begin{tabular}{l c c c c c c}
\toprule
Method & \makebox[4.8em][c]{Avg} & APEX & WSB & WorkBuddy & SB-2 & JobBench \\
\cmidrule(lr){2-2}\cmidrule(lr){3-7}
\rowcolor[HTML]{E8F0FE} \multicolumn{7}{l}{\textbf{Gemini 3.5 Flash}} \\
\rowcolor{gray!8} Single rollout & \makebox[4.8em][c]{47.2} & \cm{48.1} & \cm{56.5} & \cm{73.2} & \cm{27.3} & \cm{30.9} \\
Majority voting & \makebox[4.8em][c]{47.5} & \cs{47.3}{0.4} & \cs{57.0}{1.8} & \cs{74.0}{1.0} & \cs{27.0}{1.3} & \cs{32.4}{0.5} \\
Best-of-$N$ with judge & \makebox[4.8em][c]{47.3} & \cs{47.2}{0.4} & \cs{57.0}{2.1} & \cs{73.6}{1.2} & \cs{27.7}{0.3} & \cs{30.9}{0.9} \\
Pairwise tournament & \makebox[4.8em][c]{49.1} & \cs{49.2}{1.4} & \cs{59.6}{1.0} & \cs{74.8}{0.8} & \cs{30.4}{0.7} & \cs{31.4}{1.5} \\
LLM-as-a-Verifier & \makebox[4.8em][c]{49.5} & \cs{49.8}{0.5} & \cs{60.4}{0.2} & \cs{75.3}{0.5} & \cs{30.2}{0.5} & \cs{31.9}{1.2} \\
Agentic verifier (env. access) & \makebox[4.8em][c]{49.4} & \cs{49.0}{0.8} & \cs{60.1}{1.3} & \cs{75.1}{0.9} & \cs{30.9}{1.1} & \cs{31.8}{1.4} \\
\rowcolor{oursbg} \harnessname{} (select) & \makebox[4.8em][c]{\textbf{51.6}} & \cs{\textbf{53.3}}{1.2} & \cs{\textbf{62.0}}{0.9} & \cs{\textbf{77.8}}{1.4} & \cs{\textbf{31.5}}{2.4} & \cs{\textbf{33.5}}{1.8} \\
\rowcolor{oursbg} \quad\textit{gain over single rollout} & \makebox[4.8em][c]{\color{gainpos}$+4.4$} & \cg{{\color{gainpos}$+5.2$}} & \cg{{\color{gainpos}$+5.5$}} & \cg{{\color{gainpos}$+4.6$}} & \cg{{\color{gainpos}$+4.2$}} & \cg{{\color{gainpos}$+2.6$}} \\
\cmidrule(lr){1-7}
\multicolumn{7}{l}{\textit{Deployed in existing agent CLIs}} \\
\harnessname{} in Gemini CLI & \makebox[4.8em][c]{51.5} & \cs{51.3}{1.1} & \cs{63.2}{1.1} & \cs{77.6}{0.8} & \cs{31.5}{1.0} & \cs{33.8}{1.2} \\
\harnessname{} in Codex & \makebox[4.8em][c]{51.4} & \cs{51.5}{1.1} & \cs{63.8}{1.1} & \cs{75.8}{1.1} & \cs{30.5}{1.1} & \cs{35.6}{1.3} \\
\cmidrule(lr){1-7}
\rowcolor{gray!8} Selection oracle & \makebox[4.8em][c]{62.7} & \cm{67.3} & \cm{70.3} & \cm{87.7} & \cm{37.4} & \cm{51.0} \\
\cmidrule(lr){1-7}
\multicolumn{7}{l}{\textit{With revision}} \\
Aggregation over the pool & \makebox[4.8em][c]{49.9} & \cs{49.6}{1.0} & \cs{60.9}{1.5} & \cs{74.4}{1.1} & \cs{29.0}{1.2} & \cs{35.6}{1.6} \\
Agentic verifier + revision & \makebox[4.8em][c]{$50.7$} & \cs{50.5}{1.0} & \cs{62.0}{1.5} & \cs{76.0}{1.0} & \cs{31.0}{1.2} & \cs{34.0}{1.5} \\
\rowcolor{oursbg} \harnessname{} & \makebox[4.8em][c]{\textbf{53.4}} & \cs{\textbf{54.8}}{1.0} & \cs{\textbf{65.4}}{1.0} & \cs{\textbf{78.2}}{0.2} & \cs{\textbf{32.2}}{0.6} & \cs{\textbf{36.2}}{1.1} \\
\rowcolor{oursbg} \quad\textit{gain over single rollout} & \makebox[4.8em][c]{\color{gainpos}$+6.2$} & \cg{{\color{gainpos}$+6.7$}} & \cg{{\color{gainpos}$+8.9$}} & \cg{{\color{gainpos}$+5.0$}} & \cg{{\color{gainpos}$+4.9$}} & \cg{{\color{gainpos}$+5.3$}} \\
\midrule
\rowcolor[HTML]{E8F0FE} \multicolumn{7}{l}{\textbf{Claude Opus 4.8}} \\
\rowcolor{gray!8} Single rollout & \makebox[4.8em][c]{49.6} & \cm{35.8} & \cm{60.5} & \cm{79.0} & \cm{30.1} & \cm{42.8} \\
Majority voting & \makebox[4.8em][c]{50.4} & \cs{38.0}{0.2} & \cs{62.0}{0.7} & \cs{79.3}{0.7} & \cs{29.5}{0.5} & \cs{43.3}{1.5} \\
Best-of-$N$ with judge & \makebox[4.8em][c]{50.6} & \cs{38.4}{0.4} & \cs{62.8}{0.3} & \cs{79.2}{1.1} & \cs{30.5}{0.0} & \cs{42.3}{1.9} \\
Pairwise tournament & \makebox[4.8em][c]{51.3} & \cs{38.4}{1.1} & \cs{63.1}{0.7} & \cs{80.5}{1.4} & \cs{31.4}{0.7} & \cs{42.9}{0.9} \\
LLM-as-a-Verifier & \makebox[4.8em][c]{51.2} & \cs{36.7}{0.8} & \cs{62.4}{0.2} & \cs{81.7}{0.3} & \cs{30.2}{0.3} & \cs{45.0}{1.8} \\
Agentic verifier (env. access) & \makebox[4.8em][c]{51.6} & \cs{38.1}{0.9} & \cs{63.4}{0.6} & \cs{80.9}{0.7} & \cs{31.8}{0.8} & \cs{43.9}{1.3} \\
\rowcolor{oursbg} \harnessname{} (select) & \makebox[4.8em][c]{\textbf{53.7}} & \cs{\textbf{41.2}}{0.7} & \cs{\textbf{65.0}}{0.5} & \cs{\textbf{83.0}}{0.4} & \cs{\textbf{33.4}}{1.2} & \cs{\textbf{46.1}}{0.8} \\
\rowcolor{oursbg} \quad\textit{gain over single rollout} & \makebox[4.8em][c]{\color{gainpos}$+4.1$} & \cg{{\color{gainpos}$+5.4$}} & \cg{{\color{gainpos}$+4.5$}} & \cg{{\color{gainpos}$+4.0$}} & \cg{{\color{gainpos}$+3.3$}} & \cg{{\color{gainpos}$+3.3$}} \\
\cmidrule(lr){1-7}
\multicolumn{7}{l}{\textit{Deployed in existing agent CLIs}} \\
\harnessname{} in Claude Code & \makebox[4.8em][c]{51.7} & \cs{39.0}{1.1} & \cs{64.9}{1.3} & \cs{81.5}{0.9} & \cs{30.9}{1.0} & \cs{42.4}{1.4} \\
\harnessname{} in Codex & \makebox[4.8em][c]{52.2} & \cs{38.4}{1.1} & \cs{63.9}{1.2} & \cs{81.8}{0.9} & \cs{33.1}{1.0} & \cs{43.6}{1.5} \\
\cmidrule(lr){1-7}
\rowcolor{gray!8} Selection oracle & \makebox[4.8em][c]{63.1} & \cm{53.8} & \cm{72.4} & \cm{89.4} & \cm{39.6} & \cm{60.4} \\
\cmidrule(lr){1-7}
\multicolumn{7}{l}{\textit{With revision}} \\
Aggregation over the pool & \makebox[4.8em][c]{52.3} & \cs{40.2}{1.1} & \cs{63.5}{0.8} & \cs{80.8}{0.9} & \cs{30.9}{0.7} & \cs{45.9}{1.4} \\
Agentic verifier + revision & \makebox[4.8em][c]{$53.1$} & \cs{42.0}{1.0} & \cs{64.5}{0.8} & \cs{82.0}{0.7} & \cs{32.0}{1.0} & \cs{45.0}{1.3} \\
\rowcolor{oursbg} \harnessname{} & \makebox[4.8em][c]{\textbf{56.1}} & \cs{\textbf{47.5}}{0.2} & \cs{\textbf{67.3}}{0.2} & \cs{\textbf{83.7}}{0.7} & \cs{\textbf{34.3}}{0.2} & \cs{\textbf{47.6}}{0.5} \\
\rowcolor{oursbg} \quad\textit{gain over single rollout} & \makebox[4.8em][c]{\color{gainpos}$+6.4$} & \cg{{\color{gainpos}$+11.7$}} & \cg{{\color{gainpos}$+6.8$}} & \cg{{\color{gainpos}$+4.7$}} & \cg{{\color{gainpos}$+4.2$}} & \cg{{\color{gainpos}$+4.8$}} \\
\bottomrule
\end{tabular}
\end{table}

\subsection{Results}
\noindent\textbf{\harnessname{} outperforms all selection baselines across both models.} Table~\ref{tab:main} shows that it improves the single-rollout average by 4.4 points with Flash and 4.1 points with Opus, exceeds every baseline in all ten model--benchmark settings, and gains more than the cross-seed spread in every cell. Majority voting and best-of-$N$ judging yield only modest average gains on these long-horizon workspace tasks. The consistent advantage over methods that judge only the candidate pool supports the value of checking claims against environmental evidence, and the agentic verifier with the same environment access but no protocol or skills recovers only about half of the harness's gain, so access alone is not sufficient.

\noindent\textbf{The gain transfers to existing agent harnesses.} Running the protocol inside Gemini CLI, Claude Code, or Codex instead of our own implementation preserves most of the selection improvement: with Flash, both CLIs match our implementation within two points on every benchmark, and with Opus the CLIs recover most of the gain on Workspace-Bench Lite and WorkBuddy Bench and about half of it on APEX-Agents, but fall back to the single-rollout level on JobBench, as does Claude Code on SpreadsheetBench~2. This result shows that the verification protocol and skills are not tied to one agent implementation; Appendix~\ref{app:runtime} discusses why we still provide a dedicated one.

\noindent\textbf{Evidence-backed revision adds value beyond selection.} Applying the revision plan improves all ten model--benchmark settings at the cost of one editing pass per task (Appendix~\ref{app:cost}), and raises the average gain over a single rollout to 6.2 points with Flash and 6.4 points with Opus, with the largest added gain on APEX-Agents with Opus and on Workspace-Bench Lite with Flash. Aggregating the pool into a new artifact without evidence checks stays below the harness on every benchmark, so the revision gain comes from the evidence records rather than from rewriting itself.

%% file: sections/06-self-evolving.tex
\section{VeriHarness Can Self-Evolve}
\label{sec:self-evolving}

\begin{figure}[t]
\centering
\includegraphics[width=\linewidth]{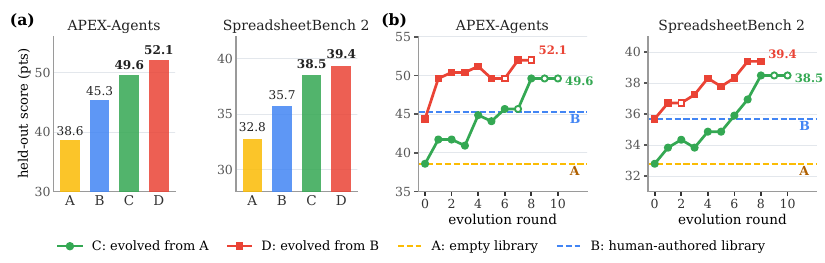}
\caption{\textbf{Evolved libraries improve held-out scores.} (a)~Held-out scores of the four libraries. A: empty; B: human-authored; C: evolved from A; D: evolved from B. (b)~Held-out scores recorded each round for C and D, with A and B as dashed lines; end labels give the final-library scores of (a). Filled markers after round 0 indicate a new library retained using development scores; hollow markers indicate that the current library is unchanged.}
\label{fig:evolve}
\end{figure}

In this section, we study whether the verification skills can accumulate from failure feedback. The skill library $\mathcal{S}$ supplies reusable knowledge of what to check and how to revise an artifact. Section~\ref{sec:experiments} uses a human-authored library. Here we let the same verifier model improve the library from development-task failures, starting either from an empty library or from the human-authored library. The model, verification protocol $\Pi$, and general tools remain fixed.

\noindent\textbf{The learning loop.} We split the eligible tasks of each benchmark into a development set and a held-out test set at a ratio of about $3{:}1$. In each round, the verifier model reads development failures together with per-item grader feedback and proposes candidate libraries. The candidate with the highest development score replaces the current library if it matches or exceeds that library's score. Held-out scores are recorded each round but never used for selection, and verification on held-out tasks sees no reference answers or grading rubrics. Appendix~\ref{app:evolve} details the split, the failure packages, the filtering of task-specific content, and the selection rule.

\noindent\textbf{Four conditions.} We compare \textbf{A}, an empty library; \textbf{B}, the frozen human-authored library; \textbf{C}, the final library evolved from A; and \textbf{D}, the final library evolved from B. All four use the full harness, including evidence-backed revision, with Claude~Opus~4.8 on APEX-Agents and SpreadsheetBench~2. These experiments use held-out subsets, so their scores are distinct from the full-benchmark results in Table~\ref{tab:main}. A versus C measures learning from failures, B versus C compares learned and human-authored skills, and B versus D measures improvement beyond human initialization.

\noindent\textbf{Skills learned from empty outperform the human-authored library.} Relative to A, C improves the held-out score by 11.0 points on APEX-Agents and 5.7 points on SpreadsheetBench~2 (Figure~\ref{fig:evolve}a). C also exceeds B on both benchmarks. With the model and protocol fixed, these comparisons show that failure feedback can improve the reusable skills that guide checking and revision.

\noindent\textbf{Human initialization improves final performance and accelerates progress.} Evolving from B yields the highest final score on both benchmarks, adding 6.8 points on APEX-Agents and 3.7 points on SpreadsheetBench~2 over B. Figure~\ref{fig:evolve}b shows that D reaches C's final trajectory score earlier on both benchmarks, with a larger reduction in rounds on APEX-Agents. Appendix~\ref{app:evolve} gives the round-by-round comparison and distinguishes progress in rounds from compute efficiency.

\noindent\textbf{Examples of learned skills.} The learned libraries give the verifier specific instructions for checking an artifact. An APEX-Agents skill instructs it to list every item the task requests, compare how the candidate answers address each item, and check the supporting evidence before revising the answer. A SpreadsheetBench~2 skill instructs it to read the period in a growth-rate label and check that the formula uses the corresponding start and end columns. These examples illustrate the reusable checking procedures added through evolution. Appendix~\ref{app:evolve} gives further examples.

\FloatBarrier

%% file: sections/07-analysis.tex
\section{Analysis}
\label{sec:analysis}

\begin{wrapfigure}{r}{0.30\linewidth}
\vspace{-12pt}
\centering
\includegraphics[width=\linewidth]{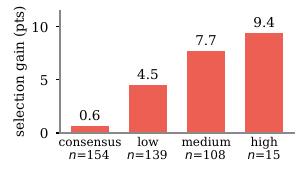}
\caption{\textbf{Selection gains where the pool disagrees.} Selection score minus pool mean on APEX-Agents, Claude~Opus~4.8, by the claim entropy of the pool.}
\label{fig:gain-by-entropy}
\vspace{-10pt}
\end{wrapfigure}
We ask four questions about the harness: which parts of the harness carry the gain, on which tasks selection gains, what happens when the verifier faces a unanimous pool, and how the evolved libraries differ from the human-authored library.

\noindent\textbf{The disagreement resolver and the consensus challenger recover different errors.} Table~\ref{tab:branches} in Appendix~\ref{app:ablation} runs the full harness with one part removed. With Claude~Opus~4.8 the resolver alone raises the five-benchmark average from 49.6 to 54.3 and the challenger alone to 52.6, and the two together reach 56.1, so their gains are complementary. The resolver's share is largest on APEX-Agents, where disputed claims are dense, and the two are closest on SpreadsheetBench~2 and JobBench, where shared errors in units, sign conventions, and omitted requirements make up more of the gap. Merging the two investigations into one context costs 0.8 points, which supports running them in separate contexts. The same ordering holds with Gemini~3.5~Flash.

\noindent\textbf{Selection gains come from the pools that disagree.} Following Figure~\ref{fig:claims}, we group the APEX-Agents tasks of Claude~Opus~4.8 by the claim entropy of their pools. On the 154 consensus pools, where the ten rollouts give the same answer, the harness's selection scores only 0.6 points above the pool mean, while on the 262 disputed pools it scores 6.1 points above. Within the disputed pools the gain grows with entropy, from 4.5 points on low-entropy pools to 7.7 on medium-entropy pools (Figure~\ref{fig:gain-by-entropy}). Section~\ref{sec:observation} observed that disagreement exposes correct alternatives, and the harness turns these alternatives into score.

\noindent\textbf{Challenging consensus generally does no harm, and its blind spot calls for more prior knowledge.} On APEX-Agents we match the challenger's verdicts on unanimous claims to the rubric items they concern (Table~\ref{tab:challenger} in Appendix~\ref{app:challenger}). The challenger is generally safe: it never refuted a claim that every rollout had right. The shared errors it left standing are mostly a matter of where it looked: in about 70\% of them it checked an intermediate result that was correct, while the error sat in a later step. The verifier tends to stop where the rollouts stopped, and telling it where else to look requires prior knowledge of how artifacts fail, which the skill library supplies.

\noindent\textbf{The evolved libraries are more specific and learn the verifier's own blind spot.} We manually compare the 95 checks in the four evolved libraries with the human-authored library: 5 restate a human check, 48 make a human check concrete with a script, a constant, or a document type, and 42 are new. Evolved skills are thus usually more specific than human-authored skills: about half of the evolved checks turn a general human check into a procedure for one kind of artifact. The new checks also cover the blind spot for consensus. The proposer reads each development failure together with the verifier's evidence record and the grader's verdicts, so it can see where the verifier confirmed a shared value that the grader marked wrong. From such cases, the libraries evolved on APEX-Agents and on SpreadsheetBench~2, in separate runs, both instruct the verifier not to treat a recomputation that matches the pool as confirmation. 

%% file: sections/08-related-work.tex
\section{Related Work}
\label{sec:related}

\noindent\textbf{Verifying agent outputs.} Verification of model outputs began as judgment: a model is prompted as a judge~\citep{zheng2023judge}, or a verifier or process reward model is trained on labeled data~\citep{cobbe2021verifiers, lightman2023verify, zhang2025genrm}. Recent work scales verification at test time through finer scores, repeated evaluation, and decomposition into checkable criteria~\citep{llmasverifier2026, multiagentverification2025, fineverify2026, asymmetricverification2025, deepverifier2026}, and a newer line lets the verifier act, by inspecting the workspace against a requirement list, gathering repository evidence, or calling tools under reinforcement learning~\citep{zhuge2024agentjudge, dockerless2026, agentvrl2026, seva2026}. Three findings limit verification by the same model: a model cannot correct its own reasoning without external feedback~\citep{huang2024selfcorrect}, its self-checks are mostly confirmatory~\citep{selfverificationdilemma2026}, and its errors are correlated with those of a judge from the same model family~\citep{goel2025alike}. The methods above train the verifier or check a single output against given criteria. \harnessname{} is training-free and uses the same model as the generator; it takes its targets from the disagreement and the consensus among rollouts, settles them with evidence from the environment, and returns a better rollout instead of a score.

\noindent\textbf{Test-time scaling for agents.} Repeated sampling raises the chance that some rollout is correct, but voting and reward models often fail to identify it~\citep{brown2024monkeys, snell2024scaling}. Studies of agents reach the same conclusion, with parallel scaling stalling at a verification gap and sequential scaling at a context ceiling~\citep{scalingttcagents2025, benchmarkttsagents2026}. Proposed remedies operate on the rollouts themselves: they vote over them, compare them in tournaments, or fuse them with an aggregating agent~\citep{ttcagenticcoding2026, wang2023selfconsistency, chen2023usc, wang2024moa, aggagent2026, samplescrutinizescale2025}. In test-time scaling, external evidence has mainly been used for code, through generated tests or distinguishing inputs~\citep{ehrlich2025codemonkeys, li2025sstar}. Debate also addresses the setting where the judge is no stronger than the debaters, but it relies on argument between models rather than evidence from the environment~\citep{irving2018debate, khan2024debate}. \harnessname{} brings evidence acquisition to tasks without executable tests, and it uses the rollouts to decide what to check.

\noindent\textbf{Harnesses and self-evolution.} The agent harness is now a recognized layer of engineering around a frozen model~\citep{anthropic2024agents, openai2026codexplatform, harnessdesign2026}, and \citet{huang2026envharness} extend it to the environment side. A recent line evolves the harness itself from execution traces, at test time or offline~\citep{tthe2026, selfharness2026, metaharness2026, jitagent2026, harnessengineering2026}, and a longer tradition grows skill and experience libraries without touching the weights~\citep{wang2023voyager, wang2024awm, ouyang2025reasoningbank, zhang2025dgm, selfevolvingsurvey2025}, with \citet{coevoskills2026} making verification itself an evolving skill. \citet{rethinkingharnesseval2026} find that evolved harnesses often fail to beat budget-matched test-time scaling and overfit when development and evaluation sets coincide. \harnessname{} evolves the verifier's skills rather than the generator's scaffold: candidates are admitted on a development set, evaluated on held-out tasks, and need no ground truth at deployment.

%% file: sections/09-conclusion.tex
\section{Conclusion}
\label{sec:conclusion}

We presented \harnessname{}, a verification harness that uses the generator's own model to resolve disagreement and challenge consensus through environmental evidence. Across five workspace benchmarks and two models, it achieves the highest selection scores among the evaluated baselines, while evidence-backed revision improves average performance over a single rollout by more than six points with each model. We further show that skills evolved from development feedback outperform the human-authored library on held-out tasks, and human initialization further improves final performance. Verification capability therefore does not have to wait for a stronger model. It can be built around the model that exists, from the structure of its own rollouts, the evidence in its environment, and its accumulated experience. Beyond test-time use, the verification record links every claim to its check, evidence, and verdict, and thus provides a claim-level signal that could supervise the generator; turning these records into training data for long-horizon agents is a natural next step.

%% file: sections/10-limitations.tex
\section*{Limitations}

\noindent\textbf{Cost of the rollout pool.} \harnessname{} verifies from a pool of rollouts, and the reported results use ten per task. This is the standard budget of test-time scaling: every baseline in Table~\ref{tab:main} receives the same ten rollouts, so the comparison is made at equal generation cost, and Appendix~\ref{app:cost} reports the cost of verification itself.

\noindent\textbf{No score for individual rollouts.} The harness delivers an artifact and a verification record, and it does not assign a scalar score to each rollout. A long-horizon artifact contains many claims that can be right or wrong independently, and the harness judges them claim by claim against evidence. Turning these judgments into a calibrated rollout-level score, for example as a reward signal, is left to future work.

\noindent\textbf{Same-model setting.} We restrict our study to a verifier that is the same model as the generator. A stronger verifier may raise the scores further, but its gain would mix the contribution of the harness with the capability gap between the two models, and separating them is outside the scope of this paper.


\noindent\textbf{Latency.} Verification is a multi-turn investigation and adds wall-clock time to a task. We target professional deliverables such as reports, workbooks, and patches, where quality matters more than response time. Deep research systems make the same trade and run for many minutes to return a better answer.

%% file: sections/appendix.tex
\newpage
\section{Claim-Level Analysis Protocol}
\label{app:claim-analysis}

This appendix describes how the claim statistics of Figure~\ref{fig:claims} are computed. The analysis uses the ten-rollout pools of Claude~Opus~4.8 on APEX-Agents and covers 434 tasks and 1{,}692 claims. It is an offline analysis that reads the benchmark's rubrics, which the verifier never sees.

\noindent\textbf{Claims and values.} Each rubric criterion names one quantity or determination that the answer should state, and we treat each criterion as one claim $p$. Claude~Opus~4.8 reads the ten answers at temperature zero and records the value $v_{i,p}$ that each rollout asserts for the criterion. Numbers count as the same value when they agree at the criterion's precision, determinations count as the same value when they state the same conclusion, and a rollout that does not address the criterion receives the value \emph{absent}. The grouped values give $\pi_p$ and $H(\pi_p)$ in Eq.~\eqref{eq:claim-entropy}, which yields 917 consensus claims and 775 disputed claims.

\noindent\textbf{Correctness.} Correctness comes from the benchmark's own grades, which mark every rollout as passing or failing each criterion. A value is correct when the majority of the rollouts asserting it pass the criterion. For consensus claims, we report how often the agreed value is correct. For disputed claims, we report how often at least one rollout passes the criterion and how often the most frequent value is correct, averaging over tied values.

\section{Why Same-Model Verification Can Improve Outputs}
\label{app:same-model-verification}

With model parameters fixed, the verifier can still operate with different information, a narrower objective, and accumulated checking experience. Table~\ref{tab:same-model-verification} explains how these differences can help it detect errors made during generation.

\begin{table}[H]
\centering
\caption{\textbf{Sources of advantage in same-model verification.}}
\label{tab:same-model-verification}
\small
\renewcommand{\arraystretch}{1.15}
\begin{tabularx}{\linewidth}{@{}p{0.22\linewidth} X@{}}
\toprule
Source & How it helps verification \\
\midrule
Exposed alternatives & Multiple rollouts make competing values and interpretations explicit. The verifier can investigate a concrete discrepancy that an individual generation attempt never encountered. \\
Focused checking & Given completed artifacts, the verifier can devote its attention to testing particular claims. Separate resolver and challenger contexts reserve attention for both disputed and shared claims. \\
Environmental evidence & A targeted source lookup or recomputation can reveal evidence omitted from the rollouts. The generator has the same tools; verification directs their use toward a specific claim whose validity is now in question. \\
Accumulated skills & Human-authored checks and procedures learned from development failures suggest what to examine, including requirements that every candidate omitted. This experience can guide later verification without changing model weights. \\
\bottomrule
\end{tabularx}
\end{table}

These advantages explain how verification can improve with a fixed model. Their value depends on finding informative checks: shared blind spots can persist when a required claim goes unexamined or the available evidence cannot settle it.

\section{Verification Procedure and Delivery Details}
\label{app:procedure}

Algorithm~\ref{alg:veriharness} summarizes the procedure described in Section~\ref{sec:method}. The program provides the workspace and model contexts; each context uses the same model to choose checks, interpret evidence, and make the decisions assigned to it.

\begin{algorithm}[htbp]
\caption{\harnessname{}}
\label{alg:veriharness}
\begin{algorithmic}[1]
\Require task $t=(s,\mathcal{E})$, rollouts $R$, model $M$, skills $\mathcal{S}$
\State expose workspace $\mathcal{W}$ containing $R$, tools $\mathcal{T}$, protocol $\Pi$, and skills $\mathcal{S}$ \Comment{program}
\State open two contexts of the verifier that run concurrently and cannot see each other \Comment{program}
\State \textbf{disagreement resolver context:} the verifier identifies the disputed claims
\While{the verifier finds a disputed claim worth a check}
  \State choose a claim $p$ and a check $\chi$ with large $\mathrm{Gain}(\chi)$ (Eq.~\eqref{eq:voi}); $e\leftarrow\mathcal{E}(\chi)$; eliminate by Eq.~\eqref{eq:eliminate}; record in $L_{\neq}$
\EndWhile
\State \textbf{consensus challenger context:} the verifier identifies the consensus claims
\While{the verifier finds a consensus claim worth a challenge}
  \State choose a check $\chi$ with high $\mathrm{Priority}(\chi)$ (Eq.~\eqref{eq:falsify}); $e\leftarrow\mathcal{E}(\chi)$; record in $L_{=}$ whether the claim holds
\EndWhile
\State open a fresh context of the verifier on $(L_{\neq},L_{=},t,R)$ \Comment{program}
\State $(\mathrm{base},\mathrm{revision\_plan},\mathrm{unresolved})\leftarrow A(L_{\neq},L_{=},t,R)$
\State $r^\star\leftarrow\mathrm{Apply}(\mathrm{base},\ \mathrm{revision\_plan},\ \mathrm{unresolved})$ \Comment{same context as adjudication}
\State $C\leftarrow(L_{\neq},\ L_{=},\ \mathrm{base},\ \mathrm{revision\_plan},\ \mathrm{unresolved})$
\State \Return $(r^\star,\ C)$
\end{algorithmic}
\end{algorithm}

\noindent\textbf{Delivery rules.} When $\mathrm{revision\_plan}$ is empty, delivery returns the selected base unchanged. Otherwise, it applies the evidence-supported changes to the base; when $\mathrm{base}=\varnothing$, it rebuilds the artifact from the inputs and evidence. Claims in $\mathrm{unresolved}$ remain marked as unsettled in the verification record. For unresolved interpretations, where the artifact format admits it, delivery presents both readings with the base's reading first; where the format has a single slot, it takes the reading preferred by adjudication.

\section{Experimental Details}
\label{app:experimental-details}

\noindent\textbf{Benchmarks.} All five benchmarks require an agent to produce an artifact from a workspace without exposing the grading rubric or reference answer. APEX-Agents~\citep{apexagents2026} contains long-horizon investment-banking, legal, and consulting tasks in file-rich simulated environments. All our experiments use version 1.0 of the benchmark. Workspace-Bench~\citep{workspacebench2026} contains realistic workspaces with large file-dependency graphs; we use its 100-task Lite subset. WorkBuddy Bench~\citep{workbuddy2026} contains tasks derived from real commits and business scenarios; we evaluate its code, office, and web domains, with 80, 50, and 70 tasks, and do not run its security domain, whose tasks consistently trigger safety alerts from the model APIs. SpreadsheetBench~2~\citep{spreadsheetbench2} evaluates financial modeling, template completion, debugging, and visualization in large workbooks. JobBench~\citep{jobbench2026} covers 35 white-collar occupations with heterogeneous reference files. 

\noindent\textbf{Models and rollout pools.} We use Gemini~3.5~Flash and Claude~Opus~4.8 with the thinking level set to high. For each benchmark and model, the generator runs in the benchmark's own agent scaffold with its standard tools and instructions and produces $N=10$ rollouts per task. The scaffolds are the APEX-Agents agent runner with its tool servers, opencode for JobBench, the SpreadsheetBench~2 agent with its shell and spreadsheet-viewing tools, and Claude Code for WorkBuddy Bench and Workspace-Bench Lite. Both models use the same scaffold on each benchmark. Each rollout includes its artifact, final workspace state, and trace. We freeze these pools before verification, so every method operates on the same candidates.

\noindent\textbf{Harness runtime.} Our runtime is a thin layer over the open-source pi coding agent (MIT license), which provides the agent loop, tool calling, and session records; Claude is served through a local LiteLLM proxy. The verifier has file reading, search, and shell tools. Rollouts, task inputs, and the environment are mounted read-only in a sandbox with no network access; only the output directory is writable. Spreadsheets are also rendered as cell-level text for comparison, and evidence skills cover spreadsheets, documents, presentations, PDFs, document bundles, and code patches (Appendix~\ref{app:tools-skills}). The two investigations run concurrently in separate contexts of the same model, and neither reads the other's record. Adjudication opens a fresh context that receives the two evidence records, and delivery continues in the adjudication context. The CLI rows of Table~\ref{tab:main} run the same instruction and skills inside Gemini CLI 0.46.0, Claude Code 2.1.246, and Codex 0.149.1, using each CLI's own tools and sandbox in place of our runtime.

\noindent\textbf{Grading protocol.} Each benchmark's own grader scores the delivered artifacts; the verifier never accesses that grader. APEX-Agents and JobBench grade each rubric item with an LLM judge, for which we use Gemini~3~Flash. Workspace-Bench Lite uses its agent-as-a-judge, a Claude Code agent with Claude~Opus~4.8. WorkBuddy Bench uses its composite verifier, which applies tests and rules in the code domain, rules and an LLM judge (Gemini~3.5~Flash) in the office domain, and the benchmark's judge pipeline in the web domain. SpreadsheetBench~2 recalculates the delivered workbook and compares it cell by cell with the reference workbook. When the grader is a judge model, the base and delivered rollouts are scored in the same round. We report the archived base score plus their same-round difference, which controls for judge drift between rounds.

\noindent\textbf{Baselines.} Every method receives the same $N=10$ rollouts and model. The \emph{single rollout} score is the pool mean. \emph{Majority voting} selects the rollout most consistent with the others, extending self-consistency~\citep{wang2023selfconsistency} to open-ended artifacts. \emph{Best-of-$N$ with a judge} scores all rollouts against the task description in one call. The \emph{pairwise tournament} repeatedly compares two rollouts and advances the winner. \emph{LLM-as-a-Verifier}~\citep{llmasverifier2026} decomposes the task into criteria and ranks candidates through per-criterion pairwise comparisons. The \emph{agentic verifier (env.\ access)} runs the best-of-$N$ judge as a tool-using agent in the same sandbox as our verifier, with the same file, search, and shell tools, but with a single scoring instruction and no protocol or skills. The \emph{selection oracle} selects the rollout with the highest grader score for each task and serves only as an upper bound on selection from the fixed pool. For the full harness, \emph{aggregation over the pool} asks the model to write a new artifact from the $N$ rollouts and the task description without environment access, following mixture-of-agents aggregation~\citep{wang2024moa, aggagent2026}. The \emph{agentic verifier + revision} extends the agentic verifier to artifact editing: in a single context, it checks the candidates against the environment, chooses the best starting artifact, and corrects the errors it finds.

\noindent\textbf{Selection and revision.} The primary comparison selects an existing rollout, isolating verification judgment from generation and allowing comparison with the oracle. The full \harnessname{} also applies the adjudicator's revision plan to the selected base. This adds one editing pass without another investigation. Methods that generate a new artifact directly from the pool, such as aggregation agents~\citep{aggagent2026}, are outside the selection comparison.

\noindent\textbf{Metrics.} Scores use each benchmark's native scale: task success rate for APEX-Agents, mean rubric score for JobBench and Workspace-Bench Lite, the task-weighted mean across WorkBuddy Bench's three domains, and headline accuracy for SpreadsheetBench~2. We report the three-seed mean, gain over the single-rollout score, and cross-seed standard deviation. The single-rollout mean and oracle are expectations over the fixed pool and require no resampling. Figure~\ref{fig:oracle-at-n} in Appendix~\ref{app:release} plots the expected selection-oracle score as a function of the pool size $N$ for both models.

\section{Released Rollout Pool}
\label{app:release}

\begin{figure}[htbp]
\centering
\includegraphics[width=\linewidth]{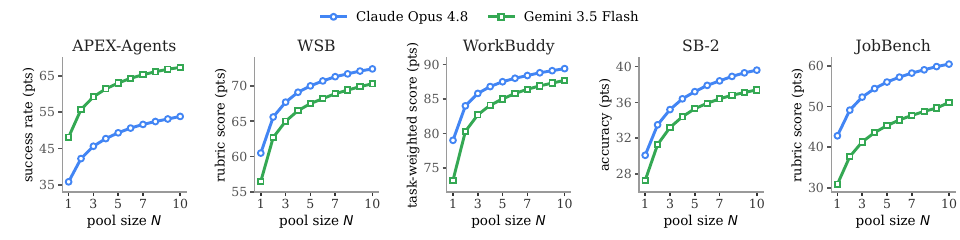}
\caption{\textbf{Selection oracle against pool size.} Expected score of the best rollout among $N$ drawn without replacement from the pool of ten, on each benchmark's own scale; $N$=1 is the single rollout and $N$=10 is the selection oracle of Table~\ref{tab:main}. The curves are computed from the grader scores in the released rollout pool.}
\label{fig:oracle-at-n}
\end{figure}

We release the rollout pools used in this paper at \url{https://huggingface.co/datasets/caiqizh/veriharness} and the verification code at \url{https://github.com/google-research/veriharness}. The release covers both models on all five benchmarks with ten rollouts per task. For each rollout, it contains the trajectory, the delivered files, and the score assigned by the benchmark's grader. The release contains no benchmark content: task descriptions, input workspaces, rubrics, and reference answers are not redistributed, and every rollout carries the official task identifier, so the inputs can be obtained from the original benchmark. Results that depend only on the pool can therefore be reproduced without running a model, including the single-rollout and selection-oracle rows of Table~\ref{tab:main} and the curves of Figure~\ref{fig:oracle-at-n}.

\section{Ablation of Harness Components}
\label{app:ablation}

Table~\ref{tab:branches} runs the full harness with only one investigation active. In the resolver-only variant, adjudication receives $L_{\neq}$ alone and consensus claims are never challenged. In the challenger-only variant, it receives $L_{=}$ alone and disputed claims are left as the base rollout states them. Both variants use the same model, tools, skills, adjudication, and delivery as \harnessname{}. The comparison applies the revision plan, because a consensus error is shared by every candidate and a challenger finding can therefore be realized only through revision. Two further variants keep both investigations: \emph{without skills} runs the protocol with an empty skill library, and \emph{single context} runs the resolver and challenger in one context, so that adjudication reads one combined record. Section~\ref{sec:analysis} discusses the contributions of the resolver and the challenger. Removing the skill library lowers the score on every benchmark, with the largest loss on APEX-Agents. Merging the two investigations into one context costs less but lowers the score on every benchmark, which supports the separate contexts of Section~\ref{sec:adjudication}. 

\begin{table}[H]
\centering
\caption{\textbf{Score of the full harness with one part disabled.} Three-seed mean on each benchmark's own scale after applying the revision plan; Avg is the unweighted mean over the five benchmarks.}
\label{tab:branches}
\footnotesize
\setlength{\tabcolsep}{5pt}
\begin{tabular}{l c c c c c c}
\toprule
Method & Avg & APEX & WSB & WorkBuddy & SB-2 & JobBench \\
\midrule
\multicolumn{7}{l}{\textbf{Gemini 3.5 Flash}} \\
Single rollout & 47.2 & 48.1 & 56.5 & 73.2 & 27.3 & 30.9 \\
Resolver only & 51.5 & 53.1 & 63.2 & 77.0 & 30.0 & 34.0 \\
Challenger only & 50.0 & 50.9 & 60.3 & 75.2 & 29.7 & 33.7 \\
\harnessname{} without skills & 51.3 & 50.9 & 63.5 & 77.1 & 30.6 & 34.4 \\
\harnessname{} in a single context & 52.6 & 53.9 & 64.5 & 77.9 & 31.3 & 35.3 \\
\harnessname{} & \textbf{53.4} & \textbf{54.8} & \textbf{65.4} & \textbf{78.2} & \textbf{32.2} & \textbf{36.2} \\
\midrule
\multicolumn{7}{l}{\textbf{Claude Opus 4.8}} \\
Single rollout & 49.6 & 35.8 & 60.5 & 79.0 & 30.1 & 42.8 \\
Resolver only & 54.3 & 44.6 & 65.5 & 82.3 & 32.9 & 46.3 \\
Challenger only & 52.6 & 39.3 & 64.0 & 80.9 & 32.6 & 46.1 \\
\harnessname{} without skills & 53.8 & 42.0 & 65.6 & 82.4 & 32.8 & 46.4 \\
\harnessname{} in a single context & 55.3 & 46.1 & 66.5 & 83.3 & 33.5 & 47.0 \\
\harnessname{} & \textbf{56.1} & \textbf{47.5} & \textbf{67.3} & \textbf{83.7} & \textbf{34.3} & \textbf{47.6} \\
\bottomrule
\end{tabular}
\end{table}

\section{Why a Dedicated Runtime}
\label{app:runtime}

\harnessname{}'s verification protocol and skills can be deployed in either our dedicated runtime or an existing agent CLI.
The CLI rows of Table~\ref{tab:main} run the protocol inside existing agent CLIs. In the terms of Eq.~\eqref{eq:harness}, a CLI keeps its own generation harness $\mathcal{H}_G$ and receives the protocol $\Pi$ and skills $\mathcal{S}$ as text in its instructions. These rows recover most of the gain, showing that the verification design transfers across runtimes. We nevertheless keep a dedicated runtime $H_V$ for three reasons.
\begin{itemize}[leftmargin=*, itemsep=2pt]
\item \emph{The program executes the protocol.} Running the two investigations in separate contexts, passing the evidence records to adjudication, and delivering the artifact with its verification record are steps that $\mathcal{H}_V$ executes explicitly. In our CLI deployments, protocol execution depends on the host agent's operating loop, including its system instructions and context management. The weaker CLI results on JobBench and SpreadsheetBench~2 with Opus suggest that the host runtime can affect how effectively the verification protocol is carried out.
\item \emph{Explicit workspace and tool configuration.} $\mathcal{W}$ and $\mathcal{T}$ expose the $N$ rollouts and the environment read-only, expose one writable output, render spreadsheets at cell level, and pass evidence records and revision plans between contexts as files. Our runtime configures these facilities explicitly and consistently across verification runs.
\item \emph{A controlled implementation.} Our runtime supports reproducibility and component ablations, as in Table~\ref{tab:branches}, while the CLI results demonstrate the portability of the verification protocol and skills.
\end{itemize}

\section{Tools and Skills}
\label{app:tools-skills}

\begin{table}[H]
\centering
\caption{\textbf{Components of the verification harness.}}
\label{tab:harness}
\footnotesize
\setlength{\tabcolsep}{6pt}
\begin{tabularx}{\linewidth}{@{}l X X@{}}
\toprule
Part & What it provides & Examples \\
\midrule
Workspace and tools $(\mathcal{W},\mathcal{T})$ & Access to the objects under check & Read a file, recompute a table, trace a source \\
Verification protocol $\Pi$ & Division of checking labor & Resolver, challenger, adjudication \\
Verification skills $\mathcal{S}$ & Reusable knowledge of what to check & Period, unit, source version, required items \\
\bottomrule
\end{tabularx}
\end{table}

\noindent\textbf{Tools.} The verifier acts through the agent's native tools: \texttt{read}, \texttt{ls}, \texttt{find}, and \texttt{grep} to look at the workspace and the rollouts, \texttt{bash} to run code, including the scripts a skill ships, and \texttt{write} and \texttt{edit}, which are restricted to the output directory, for its records and the files it delivers. This is what lets the same protocol and skills run unchanged inside Gemini CLI, Claude Code, and Codex (the CLI rows of Table~\ref{tab:main}).

\noindent\textbf{Skills.} Eighteen skills of four kinds are listed in Table~\ref{tab:skills}. A skill is a short text file, optionally with scripts, written for one kind of deliverable. The six \emph{evidence} skills tell the verifier how to read a workbook, PDF, document, deck, file bundle, or code patch as evidence rather than as extracted text, and ship the scripts for doing so. The four \emph{resolver} skills say what counts as deciding evidence when candidates of a type disagree and in which order of authority to apply it. The three \emph{challenger} skills list the ways a whole pool of that type goes wrong together, one check per entry. The five \emph{revision} skills say how the type is graded, how an unresolved claim is closed, and what must never be edited.

\begin{figure}[t]
\centering
\includegraphics[width=0.8\linewidth]{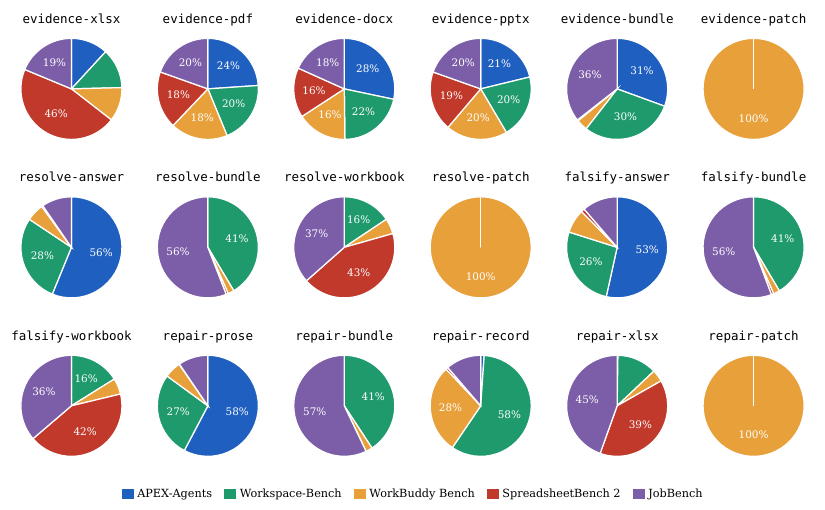}
\caption{\textbf{Where each skill was used.} For each of the eighteen skills, the number of invocations per task in the runs of Table~\ref{tab:main} in our implementation, split by benchmark with both models pooled. We report rates rather than counts so that a large benchmark does not weigh more. An invocation is the skill being loaded into a context, its file being read by the model, or one of its scripts being run.}
\label{fig:skill-use}
\end{figure}

\noindent\textbf{Where the skills are used.} Figure~\ref{fig:skill-use} shows, for each skill, how its use per task divides among the benchmarks. The document-format evidence skills are used at similar rates everywhere, except the spreadsheet skill, which SpreadsheetBench~2 draws on several times as often; the type-specific skills follow the deliverables each benchmark asks for, from written answers on APEX-Agents to patches and pages on WorkBuddy Bench.

\input{sections/tab_skills}

\FloatBarrier

\section{Self-Evolution Details}
\label{app:evolve}

\noindent\textbf{Setup and split.} The self-evolution experiments use Claude~Opus~4.8 as both verifier and library proposer. All four conditions run the complete harness, including revision, with fixed rollout pools, protocol, and general tools; evolution changes only the skill library. The APEX-Agents experiment excludes file-deliverable tasks, and the SpreadsheetBench~2 experiment covers template and financial-model tasks. Each benchmark is partitioned once into about 75\% development and 25\% held-out tasks. All ten rollouts of a task stay in one partition, and tasks that share a source workspace, source documents, or a workbook template are grouped before splitting so that related instances cannot appear on both sides. The split uses a fixed seed, balances task categories as far as the grouping permits, and is shared by conditions A--D throughout. Figure~\ref{fig:evolve} reports held-out scores under this setup, whereas Table~\ref{tab:main} reports full-benchmark pools, so condition B need not match its full-harness row.

\noindent\textbf{Information boundary.} The human-authored library is written from general document-handling knowledge and development failures only, then frozen; its authors do not inspect held-out artifacts, traces, or grades. Human-authored and evolved skills contain reusable procedures rather than task identifiers or reference answers. Held-out grades are visible only to the evaluation and are never returned to the proposer or used to change prompts, skills, candidate counts, or stopping decisions. The verification run itself never sees reference answers or grading rubrics, so evolution uses development supervision while preserving the verification-time boundary of Section~\ref{sec:setup}.

\noindent\textbf{Learning loop.} A development failure package contains the task and workspace, the fixed candidate rollouts, the harness's evidence records and revision plan, the delivered artifact, and the grader's per-item verdicts. From these packages the proposer generates $K$ candidate libraries per round, with $K=3$ at first and $K=5$ later; candidates may include scripts, which run under a ten-minute limit. An automatic filter rejects candidates that contain task identifiers, benchmark names, development filenames, or reference-answer values. Each candidate and the current library are evaluated on the same development tasks and pools; the candidate with the largest nonnegative paired gain replaces the current library, and the library is kept when every candidate has a negative gain. The held-out score is recorded after this decision, so a retained update can lower it. The schedules of ten rounds for C and eight for D were fixed before any held-out trajectory was inspected, and the final libraries are those retained at the end of the schedule rather than checkpoints chosen by held-out score.

\noindent\textbf{Round-by-round comparison.} On APEX-Agents, C first reaches its final trajectory score of 49.6 at round 8, and D reaches that score at round 1. On SpreadsheetBench~2, C first reaches 38.5 at round 8, while D first exceeds it at round 7 with 39.4, so the reduction in rounds depends on the benchmark. Both trajectories include held-out regressions, consistent with selection on development scores, and candidate counts change across rounds, so round counts alone do not establish a compute or elapsed-time advantage. The figures report point estimates and do not establish statistical significance.

\noindent\textbf{What the learned skills contain.} The following procedures illustrate the learned libraries:
\begin{itemize}[leftmargin=*, itemsep=2pt]
\item \textbf{Requirement coverage.} For a written answer, enumerate required figures, list members, and elements of a governing rule. Compare the value supplied by each rollout for each item, then check the supporting evidence before accepting or revising the base artifact's value.
\item \textbf{Normalization.} For a normalized financial line, identify each named nonrecurring item and check that the adjustment reverses its contribution under the workbook's conventions.
\item \textbf{Period alignment.} For a growth rate, identify the start and end columns from the label and check that the formula uses that interval.
\item \textbf{Sign conventions.} For a line labelled as a deduction, check how the workbook stores and applies the sign before changing a value or formula.
\end{itemize}
Because evaluation scores the delivered artifact, the gains measure the combined effect of the learned skills on checking, adjudication, and delivery. The held-out comparisons show that the learned libraries help on unseen tasks within the evaluated splits; generalization beyond these task distributions, and improvement without development grader feedback, remain to be established.

\section{Why Shared Errors Survive}
\label{app:challenger}

This appendix supports the analysis of the consensus challenger in Section~\ref{sec:analysis}. On APEX-Agents with Claude~Opus~4.8, we match each row of the challenger's evidence record that concerns a claim shared by all ten rollouts to the rubric items it concerns, using the same model to read the record and the rubric. The archived per-item grades identify the claims on which every rollout was wrong and the challenger upheld the shared value. Table~\ref{tab:challenger} classifies these surviving errors by the relation between the claim the challenger tested and the item the rubric grades.

\begin{table}[H]
\centering
\caption{\textbf{Why shared errors survive the challenger.} APEX-Agents, Claude~Opus~4.8; share of the unanimous errors that the challenger upheld.}
\label{tab:challenger}
\footnotesize
\setlength{\tabcolsep}{6pt}
\begin{tabular}{l r}
\toprule
Relation between the tested claim and the graded item & Share \\
\midrule
The tested claim is an input or intermediate quantity; the graded quantity lies downstream & 70\% \\
Same quantity, different definition, period, or source version & 14\% \\
The challenger confirmed the wrong value of the graded quantity & 12\% \\
Omission or mapping noise & 4\% \\
\bottomrule
\end{tabular}
\end{table}

\section{Compute Cost}
\label{app:cost}

\begin{figure}[htbp]
\centering
\includegraphics[width=\linewidth]{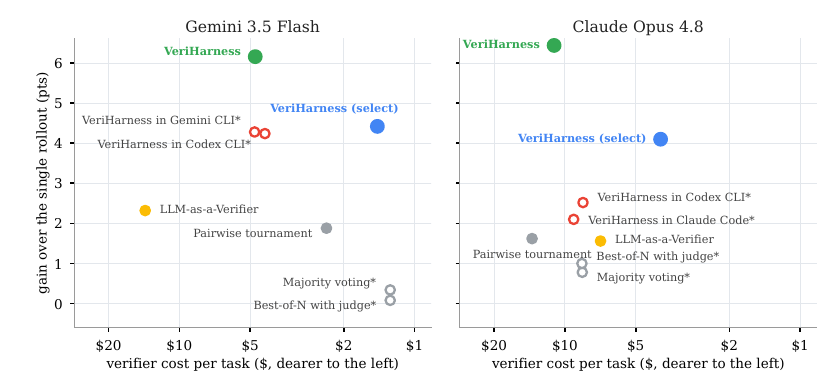}
\caption{\textbf{Cost against gain.} Compute per task at list price with cached input at the cache-read rate, on a log scale with cost decreasing to the right, against the gain over the single rollout, the five-benchmark mean of Table~\ref{tab:main}. Hollow markers with an asterisk are estimates.}
\label{fig:pareto}
\end{figure}

This section compares the compute cost of the verification methods under a single accounting. Figure~\ref{fig:pareto} plots each method's compute per task against its gain over the single rollout for both models. \harnessname{} lies on the cost--gain frontier for both models, with and without revision. With Gemini 3.5 Flash, selection costs \$1.44 per task, a tenth of the cost of LLM-as-a-Verifier at twice the gain, and less than the pairwise tournament. With Claude Opus 4.8, selection costs \$3.92, about half the cost of LLM-as-a-Verifier, majority voting, and best-of-$N$, whose ten-rollout prompts are expensive at Opus prices, and a quarter of the cost of the tournament. Applying the revision plan costs about three times as much as selection and yields the largest gain in the figure. Running the same harness inside an agent CLI reaches the same gain with Flash at three times the cost of our runtime, and a lower gain with Opus at two to three times the cost. The harness is inexpensive because its calls are incremental turns over one workspace, so 86 to 91\% of its input tokens are served from cache. When cached tokens are charged at the full input price, selection costs \$5.40 with Flash and \$13.06 with Opus, and the harness remains on the frontier, because the baselines re-read whole rollouts on every call and do not benefit from the cache.

\noindent\textbf{Metering.} Costs are computed from the token counts of the runs at list price, with cached input charged at the cache-read rate. Gains are the five-benchmark means of Table~\ref{tab:main}. LLM-as-a-Verifier is run with four evaluations per comparison. Majority voting, best-of-$N$, and the CLI runs are estimates from their call counts and context sizes rather than metered runs and are shown with hollow markers and an asterisk; the aggregation baseline calls its own client and could not be metered.

%% file: sections/tab_skills.tex
\begin{table}[htbp]
\centering
\caption{\textbf{The skill library.} The eighteen skills of the human-authored library, the deliverable each is written for, what it supplies, and the scripts it ships.}
\label{tab:skills}
\footnotesize
\setlength{\tabcolsep}{4pt}
\renewcommand{\arraystretch}{1.1}
\begin{tabularx}{\linewidth}{@{}l l >{\raggedright\arraybackslash}X >{\raggedright\arraybackslash}X@{}}
\toprule
Skill & Deliverable & What it supplies & Scripts \\
\midrule
\texttt{evidence-xlsx} & any & formulas beside cached values, workbook diffs, rendered sheets & \texttt{xlsx\_dump}, \texttt{xlsx\_diff}, \texttt{xlsx\_gaps}, \texttt{xlsx\_forks}, \texttt{xlsx\_render} \\
\texttt{evidence-pdf} & any & tables with their columns, words with coordinates, rendered pages & \texttt{pdf\_render}, \texttt{pdf\_tables}, \texttt{pdf\_words} \\
\texttt{evidence-docx} & any & text in reading order, tables, tracked changes and comments & \texttt{docx\_changes}, \texttt{docx\_text} \\
\texttt{evidence-pptx} & any & shapes in slide order, tables, charts, notes, rendered slides & \texttt{pptx\_render}, \texttt{pptx\_text} \\
\texttt{evidence-bundle} & file bundle & inventory of what each candidate delivered and states & \texttt{bundle\_inventory} \\
\texttt{evidence-patch} & patch or page & build and test of every candidate patch; headless-browser probe of a delivered page & \texttt{patchlab}, \texttt{pageprobe} \\
\midrule
\texttt{resolve-answer} & written answer & which source, provision, and question governs a split & -- \\
\texttt{resolve-bundle} & file bundle & the output contract, the basis of a count, enumerated lists, stored values & -- \\
\texttt{resolve-workbook} & workbook & order of authority: input note cells, labels, sibling formulas, task words & -- \\
\texttt{resolve-patch} & patch & executed behaviour, task words, repository idiom, project tests & -- \\
\midrule
\texttt{falsify-answer} & written answer & unopened provided documents, uncited governing clauses, statistics over the wrong set & -- \\
\texttt{falsify-bundle} & file bundle & unused stated filenames, unsurfaced breakdowns, over-broad exclusions & -- \\
\texttt{falsify-workbook} & workbook & blank labelled cells, filled unasked cells, scale, references, ranges, check rows & -- \\
\midrule
\texttt{repair-prose} & written answer & how prose is graded, what an unresolved claim permits, what never to change & -- \\
\texttt{repair-bundle} & file bundle & how office bundles are graded, coverage against the other candidates & -- \\
\texttt{repair-record} & structured record & how records are graded, filling unresolved fields, completing derived lists & -- \\
\texttt{repair-xlsx} & workbook & what in-place edits break, recalculation, proof that edits touched only their targets & \texttt{xlsx\_recalc} \\
\texttt{repair-patch} & patch & edit the tree, never the patch text; the patch must apply, reproduce, and pass what the base passed & -- \\
\bottomrule
\end{tabularx}
\end{table}